\documentclass[lettersize,journal,twoside]{IEEEtran}
\usepackage{amsmath,amsfonts}
\usepackage{algorithmic}
\usepackage{algorithm}
\usepackage{array}
\usepackage[caption=false,font=normalsize,labelfont=sf,textfont=sf]{subfig}
\usepackage{textcomp}
\usepackage{stfloats}
\usepackage{url}
\usepackage{verbatim}
\usepackage{graphicx}
\usepackage{cite}
\usepackage{multirow}
\usepackage{array}
\begin{document}

\title{Transformer-Enhanced Motion Planner: Attention-Guided Sampling for State-Specific Decision Making}

\author{Lei Zhuang, Jingdong Zhao, Yuntao Li, Zichun Xu, Liangliang Zhao and Hong Liu
\thanks{*Resrach supported by the National Natural Science Foundation of China under Projects 92148203 and T2388101. (\textit{Corresponding author: Jingdong Zhao. e-mail: zhaojingdong@hit.edu.cn})}
\thanks{The authors are with the State Key Laboratory of Robotics and System, Harbin Institute of Technology, Harbin 150001, China.}}

\markboth{}%
{}

\IEEEpubid{}

\maketitle

\begin{abstract}
Sampling-based motion planning (SBMP) algorithms are renowned for their robust global search capabilities. However, the inherent randomness in their sampling mechanisms often result in inconsistent path quality and limited search efficiency. In response to these challenges, this work proposes a novel deep learning-based motion planning framework, named Transformer-Enhanced Motion Planner (TEMP), which synergizes an Environmental Information Semantic Encoder (EISE) with a Motion Planning Transformer (MPT). EISE converts environmental data into semantic environmental information (SEI), providing MPT with an enriched environmental comprehension. MPT leverages an attention mechanism to dynamically recalibrate its focus on SEI, task objectives, and historical planning data, refining the sampling node generation. To demonstrate the capabilities of TEMP, we train our model using a dataset comprised of planning results produced by the RRT*. EISE and MPT are collaboratively trained, enabling EISE to autonomously learn and extract patterns from environmental data, thereby forming semantic representations that MPT could more effectively interpret and utilize for motion planning. Subsequently, we conducted a systematic evaluation of TEMP's efficacy across diverse task dimensions, which demonstrates that TEMP achieves exceptional performance metrics and a heightened degree of generalizability compared to state-of-the-art SBMPs.
\end{abstract}


\section{Introduction}

\IEEEPARstart{M}{otion} planning remains a pivotal challenge in robotics, focusing on devising collision-free trajectories from a given start to a target using various algorithms \cite{Lavalle2006}. In recent decades, researchers have explored a multitude of approaches to tackle the intricacies of motion planning, including graph search algorithms, artificial potential field (APF), and sampling-based motion planning (SBMP). Graph search algorithms such as the Dijkstra algorithm \cite{Dijkstra1959} and A* \cite{Hart1968} ensure optimal solutions but often face challenges in high-dimensional tasks due to exponentially expanding search spaces that impair computational efficiency. Furthermore, A* and its variants \cite{Stentz1994}, \cite{Likhachev2004} depend heavily on heuristic functions, complicating their application in complex environments. The APF \cite{Khatib1986} is noted for its computational speed but prone to encountering local minima or deadlocks, particularly in environments cluttered with obstacles. In contrast, SBMPs, exemplified by the Rapidly-exploring Random Tree (RRT) \cite{Lavalle2006}, employ random sampling to circumvent the pitfalls of local optima, thereby exhibiting commendable capabilities in global search.

SBMPs adeptly overcome obstacles by leveraging collision detection mechanisms, exhibiting convincing adaptability to high-dimensional systems. The development of RRTs has inspired the creation of numerous variants, each advancing and elaborating on the foundational methodology. RRT-Connect \cite{Kuffner2000} utilizes a bidirectional search strategy to improve planning efficiency. RRT* \cite{Karaman2011} incorporates ChooseParent and Rewiring mechanisms to attain asymptotic optimality. Informed-RRT* (IRRT*) \cite{Gammell2014} mitigates the conflict between asymptotic optimality and real-time performance by constraining the sampling area. The stochastic nature of sampling in these algorithms ensures probabilistic completeness. However, uniform sampling throughout the configuration space might produce an excess of redundant nodes and potentially under-explore crucial areas necessary for efficient path discovery.

\IEEEpubidadjcol

As deep learning technologies continue to evolve, integrating traditional planning algorithms with deep learning models to expedite the planning process has become a promising approach \cite{Qureshi2018,Ichnowski2020,Xia2023,Kim2020}. A bidirectional iterative planning algorithm, MPNet \cite{Qureshi2021}, utilizes advanced neural architectures—namely ENet, an encoding network, and PNet, a planning network—to collaboratively improve the planning process through iterative neural sampling. Furthermore, harnessing a CNN model, Neural RRT* \cite{Wang2020} predicts the probability distribution of optimal paths in planning tasks, thereby facilitating non-uniform sampling in subsequent stages. Moreover, L-SBMP \cite{Ichter2019} employs an autoencoder network, a dynamics network, and a collision detection network to construct a plannable latent space, demonstrating reliable performance in high-dimensional robotic motion planning.

Transformer, a revolutionary innovation in the field of deep learning, has fundamentally altered the way sequence data is processed. At its heart, the attention mechanism of the Transformer assigns weights to each element in the sequence, efficiently capturing the interrelationships between elements, regardless of their position \cite{Vaswani2017}. Compared to Recurrent Neural Networks (RNN) \cite{Elman1990} and Long Short-Term Memory (LSTM) model \cite{Hochreiter1997},  Transformer enhances the ability to detect long-range dependencies through global perspective, thus avoiding issues of information decay. Capitalizing on its advantages, the Transformer has achieved significant improvements in various domains such as natural language processing \cite{Radford2018,Radford2019,Devlin2018} and image recognition \cite{Ramachandran2019}, \cite{Dosovitskiy2020}. Recently, researchers have also begun exploring the application of Transformer in the field of motion planning. VQ-MPT \cite{Johnson2023} utilizes the Transformer model to segment the continuous space into discrete sets and choose sampling regions, enabling planning within these strategically selected regions through conventional SBMPs.

\begin{figure}[!t]
\centering
\captionsetup[subfigure]{font=normal,labelfont={rm},textfont=rm}
\subfloat[Scenario 1, TEMP]{\includegraphics[width=0.24\textwidth]{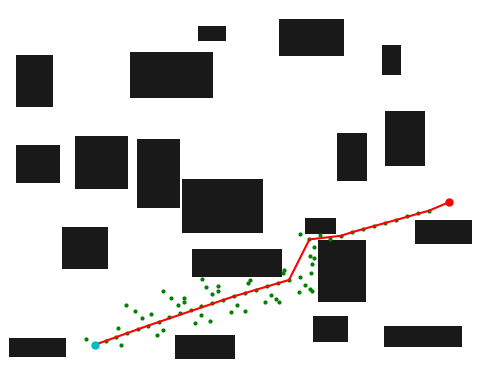}}
\hfil
\subfloat[Scenario 1, RRT*]{\includegraphics[width=0.24\textwidth]{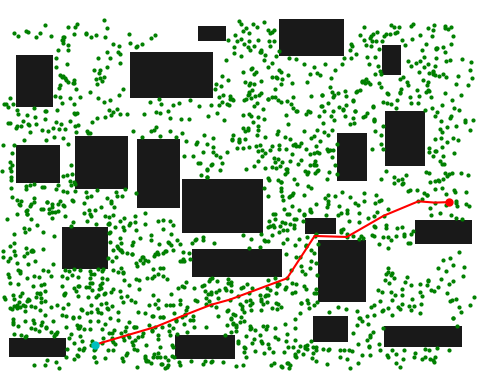}}
\\ 
\subfloat[Scenario 2, TEMP]{\includegraphics[width=0.24\textwidth]{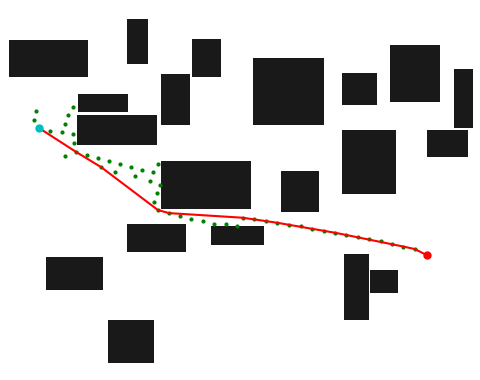}}
\hfil
\subfloat[Scenario 2, RRT*]{\includegraphics[width=0.24\textwidth]{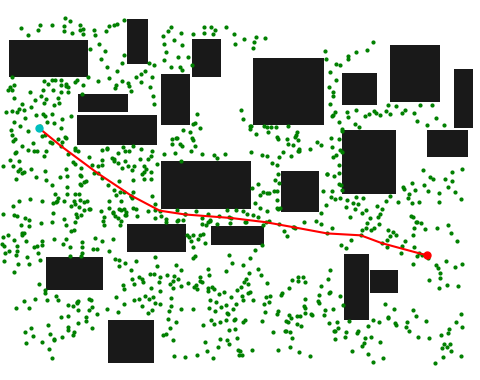}}
\caption{Performance assessment of TEMP versus RRT* in 2D planning, focusing on paths of comparable quality. The variables $t$, $N$, and $\mathcal{J}$ denote planning time, number of nodes generated, and path cost, respectively. (a) $t$ = 0.10 s, $N$ = 75, $\mathcal{J}$ = 17.78; (b) $t$ = 6.63 s, $N$ = 1599, $\mathcal{J}$ = 17.97; (c) $t$ = 0.07 s, $N$ = 51, $\mathcal{J}$ = 18.65; (d) $t$ = 4.62 s, $N$ = 902, $\mathcal{J}$ = 18.65.}
\label{Fig:FigureinIntro}
\end{figure}

When directly employing deep neural networks for sampling, current research predominantly focuses on the use of the robot's current state \cite{Qureshi2018}, \cite{Qureshi2021}, which is crucial for the next decision in the planning. Nevertheless, a comprehensive consideration of the entire path from the start to the current state could theoretically yield additional benefits, such as a better understanding of the planning process to minimize unnecessary detours. Traditional deep neural networks processing long sequence data, however, introduce new technical challenges such as vanishing gradients, exploding gradients, and issues with data retention and forgetting, which the attention mechanism of Transformer effectively mitigates \cite{Dai2019}. Although environmental information, planning task details, robot state, and path data are fed into the neural network, a considerable portion of this information might be less critical for the immediate decision step. Consequently, different levels of attention should be allocated to environmental information depending on the area's spatial characteristics; as the planning process advances, the role of start and goal on decisions needs adjustment; when a robot enters an obstacle-dense environment, emphasis on the planning destination should be minimized to prioritize escaping the predicament. These analyses illustrate the significant value of utilizing attention mechanisms to dynamically tune the focus on different parts of the sequence data throughout various stages of the planning.

The contributions of this research manifest across three dimensions. First, we develop Transformer-Enhanced Motion Planner (TEMP), an integrated framework combining Environmental Information Semantic Encoder (EISE) with Motion Planning Transformer (MPT). Drawing on the datasets compiled by RRT*, both networks participate in a collaborative training process, allowing EISE to produce Semantic Environmental Information (SEI), while MPT employs an attention mechanism to dynamically focus on SEI, task objectives, and historical planning data (HPD) during the sampling stage. Second, TEMP's attention-guided sampling notably diminishes the node count necessary for high-quality path exploration (as shown in Fig. \ref{Fig:FigureinIntro}), enhancing the sampling density in crucial areas and expediting the convergence of path cost. Third, in comparison to advanced SBMP techniques, TEMP markedly accelerates the resolution speed, achieving roughly a 10x increase over IRRT* in both 3D and 7D tasks, and it is about 24x faster than RRT* for planning in 7D. Moreover, it elevates the success rate of planning, particularly distinguishing itself in challenging, high-dimensional scenarios.

\section{Preliminaries}

\subsection{Problem Definition}

The planning space for a robot, defined as $\mathcal{X} \subseteq \mathbb{R}^n$, where $n$ indicates the dimensionality, is composed of two distinct subsets: the obstacle space $\mathcal{X}_{\text{obs}} \subset \mathcal{X}$ and the free space $\mathcal{X}_{\text{free}} = \mathcal{X} \setminus \mathcal{X}_{\text{obs}}$. Given an initial state $x_{\text{init}} \in \mathcal{X}_{\text{free}}$ and a goal region $\mathcal{X}_{\text{goal}} \subset \mathcal{X}_{\text{free}}$, the objective of motion planning is to find a feasible trajectory $\sigma = \{x_0, \ldots, x_n\}$ from $x_{\text{init}}$ to $\mathcal{X}_{\text{goal}}$. Here, $x_0 = x_{\text{init}}$, and $x_n \in \mathcal{X}_{\text{goal}}$, with the entire path residing within $\mathcal{X}_{\text{free}}$.

Searching for a collision-free trajectory $\sigma$ does not depend on the representation of $\mathcal{X}_{\text{obs}}$ but is accomplished through a collision detection algorithm. Define the obstacle space $\mathcal{W}_{\text{obs}} \subset \mathcal{W}$ and the free space $\mathcal{W}_{\text{free}} = \mathcal{W} \setminus \mathcal{W}_{\text{obs}}$ in the robot's workspace $\mathcal{W}$, where $\mathcal{W} \subseteq \mathbb{R}^m$ with $m$ as the dimension. The collision detection module, $\mathcal{D}$, checks whether any segment of the trajectory, denoted by $\overline{{x}_i, {x}_{i+1}}$, might result in a collision with the obstacles in $\mathcal{W}_{\text{obs}}$, thereby providing a guarantee of collision-free passage for the trajectory $\sigma$.

Specify $\mathcal{T}(\mathcal{X})$ as the set of all feasible trajectories within $\mathcal{X}$. For any trajectory $\sigma \in \mathcal{T}(\mathcal{X})$, the function $\mathcal{J}(\sigma)$ computes a real number representing the cost associated with that trajectory. Based on $\mathcal{J}(\sigma)$, the optimal trajectory $\sigma^*$ is identified. The complete definition of the motion planning problem is:
\begin{equation}
\label{EQ:DefMP}
\begin{aligned}
& \sigma^* = \underset{\sigma \in \mathcal{T}(\mathcal{X})}{\arg \min} \, \mathcal{J}(\sigma) \\
& \text{s.t.\quad} \sigma = \{x_0, \ldots, x_n\},\, x_0 = x_{\text{init}},\, x_n \in \mathcal{X}_{\text{goal}} \\
& \phantom{\text{s.t.\quad}} \mathcal{D}(\overline{x_i, x_{i+1}}) = \text{False},\, \forall i \in [0, n-1] \\
\end{aligned}
\end{equation}

\subsection{Sampling-Based Motion Planning}

Sampling-based motion planning explores the planning space by random sampling and constructing tree-like paths, endowing the algorithm with probabilistic completeness. Furthermore, the optimal versions of SBMP can continuously improve the path quality as the planning progresses, approaching the best solution over an infinite sample size, which bestows SBMP with asymptotic optimality.

SBMPs initiate by setting the planning start $x_\text{init}$ as the root node of the tree. In each iteration, a sample node $x_{\text{sample}}$ is randomly selected from $\mathcal{X}$, and the nearest node $x_{\text{nearest}}$ in the tree is identified as the parent node. The node $x_{\text{new}}$, generated from $x_{\text{nearest}}$ towards $x_{\text{sample}}$ by a predetermined step size,  will be added to the tree if $\mathcal{D}$ verifies that the trajectory from $x_{\text{nearest}}$ to $x_{\text{new}}$ is collision-free, and this addition continues until $\mathcal{X}_{\text{goal}}$ is reached. In optimal SBMP forms, tree expansion is governed by $\mathcal{J}(\sigma)$, prioritizing cost over distance in parent node selection after sampling. Once an initial path is established, optimal SBMPs persistently refine it by sampling and iterating based on $\mathcal{J}(\sigma)$, until reaching the algorithm's limits on iterations, runtime, or other predefined stopping conditions.

\subsection{Transformer Model}

Since its introduction, the Transformer model has become an essential method for processing sequence data. Central to this model is the attention mechanism, which enables identification of interdependencies among elements. The mathematical formulation of the attention mechanism is:
\begin{equation}
\label{EQ:Atten}
\text{Attention}(Q,K,V) = \text{softmax}\left( \frac{Q K^{T}}{\sqrt{d_{k}}} \right) V
\end{equation}
\noindent where $Q$, $K$, and $V$ represent the Query, Key, and Value vectors, respectively, while $d_k$ represents the dimensionality of the Key. In this study, we focus exclusively on the self-attention mechanism; consequently, these vectors—Query, Key, and Value—are derived from the input data and interconnected through linear mappings implemented by three trainable weight matrices: $W^Q$, $W^K$, and $W^V$. The scaling factor $d_k$ is used to normalize the dot product within the softmax function, mitigating the risk of softmax saturation and the associated issue of vanishing gradients.

Transformer processes sequential data in parallel, thus it incorporates an additional Positional Encoding module to preserve the positional information of the elements. Learnable Positional Encoding, which employs a set of positional representation parameters trained concurrently with the model, boosts the model’s capability to recognize long-range dependencies and subtle positional nuances in sequence data. Due to the influence of backpropagation, Learnable Positional Encoding directly correlates with training targets, allowing the model to discern the relationships between specific positions in the sequence and their contextual roles.

\section{Transformer-Enhanced Motion Planner}

\begin{figure*}[!t]
\centering
\includegraphics[width=\textwidth]{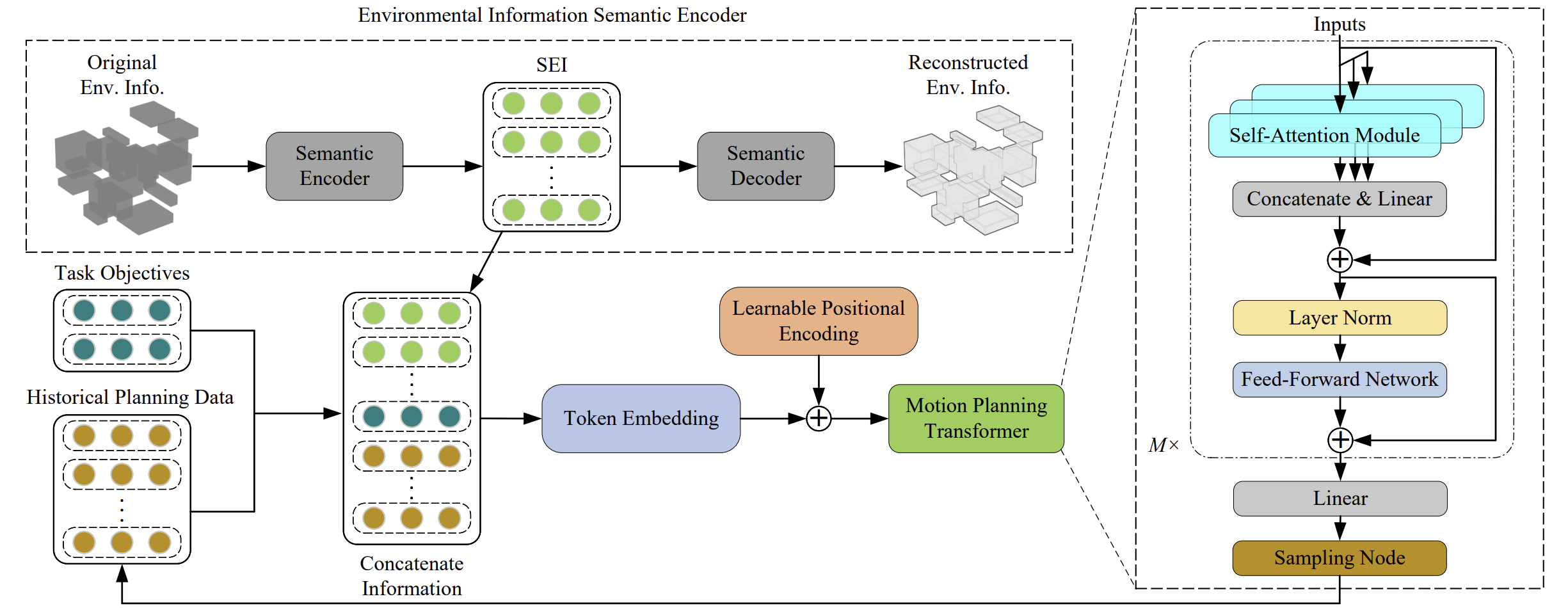}
\caption{Network architecture of the Transformer-Enhanced Motion Planner, illustrating the data flow within the system, particularly highlighting how the Environmental Information Semantic Encoder and the Motion Planning Transformer process information and contribute to generating the sampling node.}
\label{Fig:TEMPArc}
\end{figure*}

In this section, we delve into the mechanics of the proposed planning method: Transformer-Enhanced Motion Planner. The network architecture of TEMP, illustrated in Fig. ~\ref{Fig:TEMPArc}, is structured around two principal modules: the MPT and the EISE.

\subsection{Environmental Information Semantic Encoder}

The functionality of the Environmental Information Semantic Encoder is realized in multiple stages. Initially, EISE encodes the original environmental information $O$, producing a compressed representation via a fully connected layer. We employ the Rectified Linear Unit (ReLU) as the activation function to enhance the model's sparse representational capacity and accelerate convergence. Subsequently, the compressed information undergoes two distinct data flows: one passes through information integration towards the downstream planning network; the other flow is processed by a decoder, resulting in the reconstructed environmental information $\hat{O}$.

The semantic encoding capabilities of the EISE are not manually crafted but are learned through collaborative training with the MPT. Beyond merely minimizing reconstruction loss, the training objectives include adapting the EISE based on deviations in sampling nodes. Such adjustments foster the creation of semantically enriched and structured information, termed SEI $S=f_{\text{semantic}\_\text{enc}}(O)$, strengthening the MPT's ability to interpret data effectively. The loss function employed by the EISE is delineated below:
\begin{equation}
\label{EQ:EISELoss}
L_{\text{EISE}} = \lambda \cdot L_{\text{recons}}(O, \hat{O}) + \eta \cdot L_{\text{semantic}}(x_{\text{sample}}, \hat{x}_{\text{sample}})
\end{equation}
\noindent where $L_{\text{recons}}$ is the loss associated with the reconstruction of environmental information, $\hat{x}_{\text{sample}}$ denotes the model's predicted sampling node, $x_{\text{sample}}$ represents the true value of the node, and $L_{\text{semantic}}$ indicates the loss term that assesses the accuracy of semantic encoding. The hyperparameters $\lambda$ and $\eta$ are employed to modulate the relative weights of these loss components.

\subsection{Motion Planning Transformer}

The Motion Planning Transformer integrates data from the SEI $S$, task objectives $T$, and HPD $H$ within a cohesive mathematical framework, utilizing an attention mechanism to compute correlation weights among these elements. To adapt more effectively to positional cues, we employ Learnable Positional Encoding, which increases the model's flexibility and sensitivity to positional information. Given the intricate and diverse relationships between data sources, a single self-attention module may not fully capture all subtleties. Thus, we implement a multi-head self-attention mechanism, mathematically formulated as:
\begin{equation}
\label{EQ:MulHeaAtten}
\begin{aligned}
&\text{MultiHeadAttn}(Z) = \text{Concat}(A_1(Z), \ldots, A_h(Z)) W^O \\
&\text{where } A_i(Z) = \text{softmax}\left( \frac{(Z W_i^Q) (Z W_i^K)^T}{\sqrt{d_k}} \right) (Z W_i^V) \\
&\phantom{\text{where }} Z = f_{\text{emb}}(\text{Concat}(S, T, H)) + P \\
\end{aligned}
\end{equation}
\noindent where $Z$ represents the data obtained after the concatenated information from $S$, $T$, and $H$ is processed through Token Embedding $f_\text{emb}$, which maps data to a high-dimensional space to facilitate deeper feature extraction, and Positional Encoding $P$. $A_i(Z)$ reflects the output of the $i^{th}$ attention head, and $W^O$ is the weight matrix for merging outputs. $W_i^Q$, $W_i^K$, and $W_i^V$ are the query, key, and value weight matrices for the $i^{th}$ attention head, respectively.

When processing motion planning data under various environmental conditions and task directives, ensuring numerical stability is crucial for the model's rapid convergence. To this end, we incorporate Layer Normalization to standardize the features:
\begin{equation}
\label{EQ:LayerNorm}
\text{LN}(x) = \frac{x - \mu}{\nu} \cdot \alpha + \delta
\end{equation}
\noindent where $\mu$ and $\nu$ are the mean and standard deviation computed over the features of input $x$, respectively. Parameters $\alpha$ and $\delta$ are learned during training, allowing for scaling and shifting of the normalized data.

To augment the capacity for capturing intricate data patterns, the MPT processes the data through a Feed-Forward Network (FFN):
\begin{equation}
\label{EQ:FFN}
\text{FFN}(x) = \max(0, xW_1 + b_1)W_2 + b_2
\end{equation}
\noindent where $x$ denotes the data input to the FFN, $W_1$, $W_2$, $b_1$, and $b_2$ are its trainable parameters. Furthermore, to improve the model's gradient flow and ensure effective transmission of crucial features in deeper layers, we incorporate residual connections.

The Motion Planning Transformer stacks $M$ layers of the previously described network architectures to deepen the contextual analysis, expanding its ability to discern complex patterns and dependencies across vast datasets. Ultimately, the MPT maps the high-dimensional embedded data into the planning space, resulting in the generation of sampling node.

\subsection{Transformer-Enhanced Motion Planner}

The Transformer-Enhanced Motion Planner is structured into two planning processes: attention-based planning (AP) and classical planning (CP). The focus of this study, AP, is detailed in Algorithm \ref{Alg:TEMPAP}. We define the current path $\sigma'$ as the route from $x_{\text{init}}$ to the robot's current state $x_{\text{c}}$, which serves as HPD input for the MPT.

In the planning iteration phase, the MPT generates $x_{\text{sample}}$ and the Steer function calculates $x_{\text{new}}$ based on a predefined step size. If CollisionCheck returns True, the system introduces randomness to ensure TEMP's probabilistic completeness. The algorithm employs the FindNear and ChooseParent modules to identify the optimal local parent node for $x_{\text{new}}$. Subsequently, the Rewire function adjusts the surrounding connections of $x_{\text{new}}$ to elevate the path quality. Upon establishing a feasible path $\sigma$ after $n$ iterations, TEMP transitions to CP, where random sampling replaces MPT to ensure asymptotic optimality.

\begin{algorithm}[!t]
\caption{TEMP Attention-based Planning}
\label{Alg:TEMPAP}
\begin{algorithmic}[1]
\REQUIRE $O, x_{\text{init}}, \mathcal{X}_{\text{goal}}, \mathcal{J}$
\ENSURE $\sigma$
\STATE $T \gets x_{\text{init}},\,\, \sigma' \gets x_{\text{init}}$
\STATE $S \gets \text{EISE}(O, x_{\text{init}}, \mathcal{X}_{\text{goal}})$
\STATE $\sigma \gets \text{null},\,\, \mathcal{J}_{\text{best}} \gets \infty$
\FOR{$i = 1$ \TO $n$}
    \STATE $x_{\text{sample}} \gets \text{MPT}(S, \mathcal{X}_{\text{goal}}, \sigma')$
    \STATE $x_{\text{new}} \gets \text{GetNearestNode\,\&\,Steer}(T, x_{\text{sample}}, e)$
    \WHILE{$ \text{CollisionCheck}(x_{\text{new}}, x_{\text{nearest}})$}
        \STATE $x_{\text{sample}} \gets \text{RandomSample}()$
        \STATE $x_{\text{new}} \gets \text{GetNearestNode\,\&\,Steer}(T, x_{\text{sample}}, e)$
    \ENDWHILE
    \STATE $x_{\text{new}} \gets \text{FindNear\,\&\,ChooseParent}(x_{\text{new}}, T, \mathcal{J})$
    \STATE $T.\text{Append}(x_{\text{new}})$
    \STATE $\text{Rewire}(x_{\text{new}}, T, \mathcal{J})$
    \STATE $\sigma' \gets \text{UpdatePath}(x_{\text{new}}, T)$
    \IF{$ x_{\text{new}} \in \mathcal{X}_{\text{goal}}$}
        \STATE $\sigma \gets \sigma'$, $\mathcal{J}_{\text{best}} \gets \mathcal{J}(\sigma')$
        \RETURN $\sigma$
    \ENDIF
\ENDFOR
\end{algorithmic}
\end{algorithm}

\section{Simulation Results}

We performed extensive simulations to evaluate the TEMP across various tasks, including 2D and 3D point robots and a 7D manipulator using the Kuka iiwa14 model. During the evaluation, we employed two advanced SMBPs, RRT* and IRRT*, as benchmarks. The simulations were executed on a computer with a Linux OS, a 3.2 GHz Intel i9-12900KF CPU, 128 GB RAM, and an NVIDIA GeForce RTX 3090 GPU.

\begin{figure*}[!t]
\centering
\captionsetup[subfigure]{font=normal,labelfont={rm},textfont=rm}
\subfloat[Easy 2D]{\includegraphics[width=0.23\textwidth]{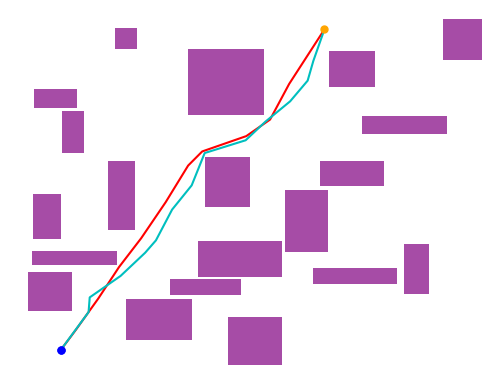}}
\hfil
\subfloat[Challenging 2D]{\includegraphics[width=0.23\textwidth]{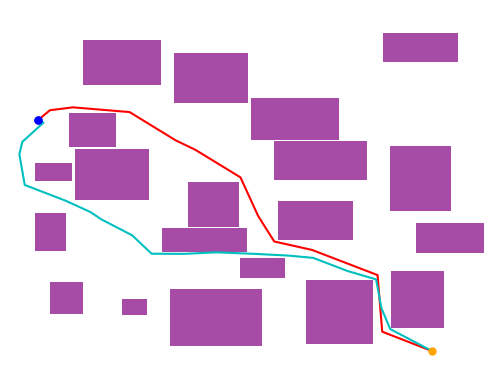}}
\hfil
\subfloat[Challenging 2D]{\includegraphics[width=0.23\textwidth]{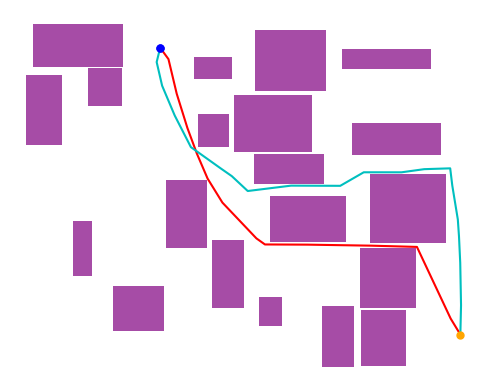}}
\hfil
\subfloat[Challenging 2D]{\includegraphics[width=0.23\textwidth]{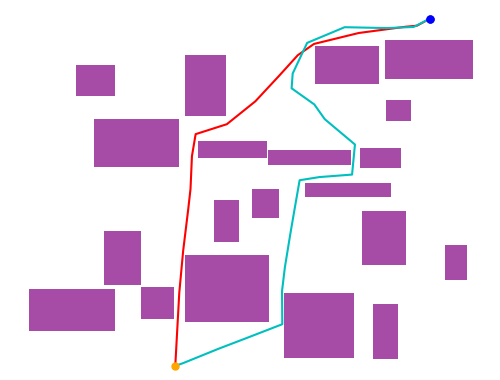}}
\\ 
\subfloat[Easy 3D]{\includegraphics[width=0.23\textwidth]{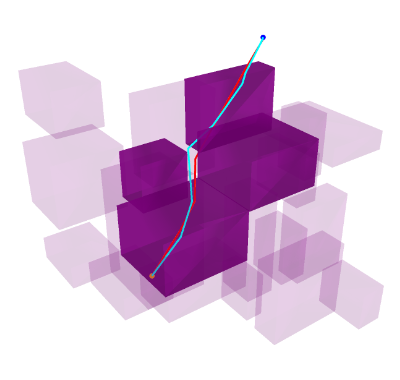}}
\hfil
\subfloat[Challenging 3D]{\includegraphics[width=0.23\textwidth]{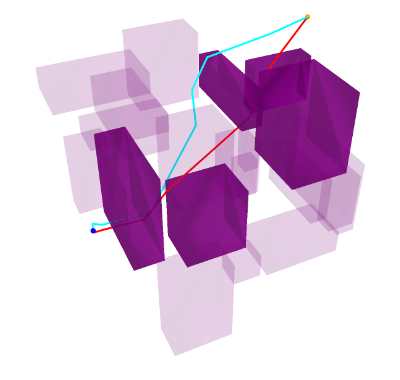}}
\hfil
\subfloat[Challenging 3D]{\includegraphics[width=0.23\textwidth]{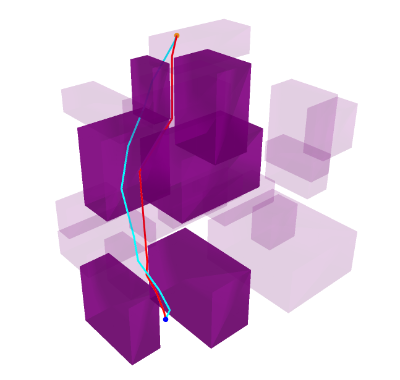}}
\hfil
\subfloat[Challenging 3D]{\includegraphics[width=0.23\textwidth]{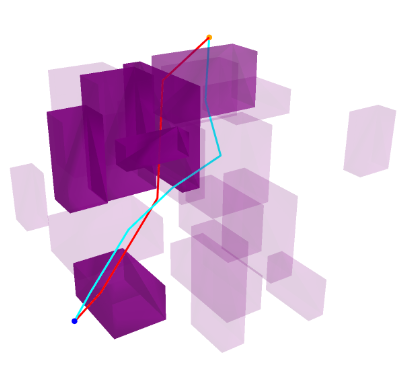}}
\caption{Comparative analysis of planning solutions between TEMP (Red) and IRRT* (Cyan) in 2D and 3D scenarios. The symbols $t_{\text{T}}$ and $t_{\text{I}}$ represent the planning times for TEMP and IRRT*, respectively; similarly, $\mathcal{J}_{\text{T}}$ and $\mathcal{J}_{\text{I}}$ indicate the path cost for each algorithm. Due to the dense distribution of obstacles in the 3D planning environments, we have rendered those obstacles that have a relatively small impact on the path planning more transparent, to improve the clarity of the displayed results. (a) $t_{\text{T}}$ = 0.10 s, $\mathcal{J}_{\text{T}}$ = 21.87, $t_{\text{I}}$ = 1.08 s, ${{\mathcal{J}}_{\text{I}}}$ = 22.19; (b) $t_{\text{T}}$ = 0.09 s, $\mathcal{J}_{\text{T}}$ = 24.73, $t_{\text{I}}$ = 3.34 s, ${{\mathcal{J}}_{\text{I}}}$ = 26.01; (c) $t_{\text{T}}$ = 0.11 s, $\mathcal{J}_{\text{T}}$ = 23.97, $t_{\text{I}}$ = 1.58 s, ${{\mathcal{J}}_{\text{I}}}$ = 27.70; (d) $t_{\text{T}}$ = 0.41 s, $\mathcal{J}_{\text{T}}$ = 25.71, $t_{\text{I}}$ = 4.50 s, ${\mathcal{J}}_{\text{I}}$ = 29.77; (e) $t_{\text{T}}$ = 0.14 s, $\mathcal{J}_{\text{T}}$ = 20.91, $t_{\text{I}}$ = 1.78 s, $\mathcal{J}_{\text{I}}$ = 21.51; (f) $t_{\text{T}}$ = 0.17 s, $\mathcal{J}_{\text{T}}$ = 24.29, $t_{\text{I}}$ = 2.17 s, $\mathcal{J}_{\text{I}}$ = 26.59; (g) $t_{\text{T}}$ = 0.29 s, $\mathcal{J}_{\text{T}}$ = 26.10, $t_{\text{I}}$ = 3.53 s, $\mathcal{J}_{\text{I}}$ = 26.91; (h) $t_{\text{T}}$ = 0.07 s, $\mathcal{J}_{\text{T}}$ = 22.02, $t_{\text{I}}$ = 3.69 s, $\mathcal{J}_{\text{I}}$ = 22.82.}
\label{Fig:PathTEMPvsIRRT}
\end{figure*}

\subsection{Networks Training and TEMP Testing Setup}

Both the EISE and MPT were developed using the PyTorch framework. We generated 80 training and 20 validation workspaces for 2D, 3D, and 7D planning tasks, respectively. In each workspace, 200 start-goal pairs were established, and paths were obtained through RRT* to form the dataset.

A collaborative training approach was implemented for the EISE and MPT. We employed the Adam optimizer, configuring it with parameters $\beta_1 = 0.9$, $\beta_2 = 0.999$ and $\epsilon = 1e-8$. The learning rate, initially set at 0.001, was reduced by a factor of 0.1 if the validation loss did not decrease by at least 10\% after ten epochs, continuing until it reached a minimum threshold of $1e-6$.

For TEMP testing, we established 100 easy and 100 challenging tasks in both 2D and 3D scenarios, based on the complexity of the planning solutions required. Additionally, 200 planning tasks were created for the 7D manipulator scenario. All testing tasks featured randomly generated obstacles and start-goal pairs, which were not included in the TEMP’s training dataset.

\subsection{Comparison of TEMP with Advanced SBMPs}

\begin{figure*}[!t]
\centering
\captionsetup[subfigure]{font=normal,labelfont={rm},textfont=rm}
\subfloat[]{\includegraphics[width=0.35\textwidth]{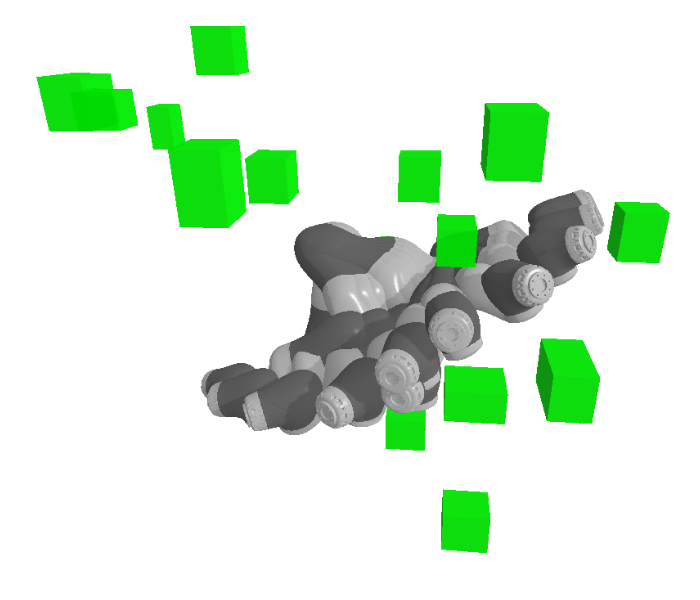}}
\hfil
\subfloat[]{\includegraphics[width=0.35\textwidth]{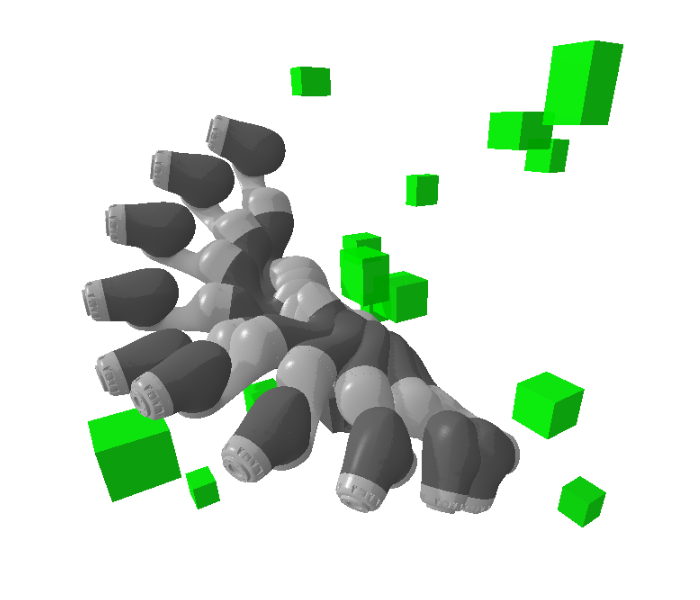}}
\caption{Partial intermediate configurations of the path generated by the TEMP in planning for a 7-DOF manipulator. (a) $t$ = 0.33 s; (b) $t$ = 0.34 s.}
\label{Fig:PathTEMPin7DOF}
\end{figure*}

We present a series of planning cases for 2D and 3D point robots in Fig. \ref{Fig:PathTEMPvsIRRT}, illustrating that TEMP can efficiently find near-optimal paths in a brief period. In contrast, IRRT* requires ten to several dozen times longer, and the quality of the paths it produces still falls short of those generated by TEMP. This difference is particularly evident in challenging tasks, as shown in Fig. \ref{Fig:PathTEMPvsIRRT}(c) and Fig. \ref{Fig:PathTEMPvsIRRT}(d), where narrow and critical areas significantly impact path solving, and TEMP explores these regions more effectively. Furthermore, Fig. \ref{Fig:PathTEMPin7DOF} displays two examples of TEMP applied to planning for a 7-DOF manipulator, showing partial intermediate configurations of planned paths. Even in environments with complex obstacle distributions, TEMP completes these planning tasks in sub-second durations, demonstrating its robust performance and reliability in handling high-dimensional motion planning.

To rigorously verify the effectiveness of the methodologies we proposed, all performance metrics of TEMP in this paper derive from its AP phase, except for the cost convergence analysis in Fig. \ref{Fig:CostvsNodes}, which inevitably encompasses the CP. Additionally, in comparing algorithm performance, we allow for a certain degree of cost tolerance for the RRT* and IRRT*:
\begin{equation}
\label{EQ:CostTol}
\begin{aligned}
& \mathcal{J}_{\text{RRT*,\,IRRT*}} \le (1+\tau) \mathcal{J}_{\text{TEMP}} \\ 
& \text{s.t.}\quad \mathcal{J} = \sum_{i=0}^{n-1} \|x_{i+1} - x_i\|_2 \\
\end{aligned}
\end{equation}
\noindent where $\tau$ represents the cost tolerance, set at 0.05. Similarly, except for the cost analysis that includes global data, we selected the best values for all other statistical metrics of RRT* and IRRT* that conform to the criteria set by Equation (\ref{EQ:CostTol}).

The average planning time (in seconds) and number of nodes for TEMP, RRT*, and IRRT* in 2D, 3D and 7D planning tasks are presented in Table \ref{Tab:ComTimeNodes}. For tasks with equivalent dimensions and difficulty, TEMP outperforms RRT* and IRRT* in both time efficiency and node utilization. Specifically, TEMP achieves solution times approximately ten times faster than the next best IRRT* in 3D and 7D scenarios, and about twenty-four times faster than RRT* in 7D. Additionally, unlike the significant increases in solution time that RRT* and IRRT* experience as task difficulty and dimensionality rise, TEMP maintains a relatively stable measure in this respect, demonstrating commendable adaptability.

\begin{figure*}[!t]
\centering
\subfloat{\includegraphics[width=0.33\textwidth]{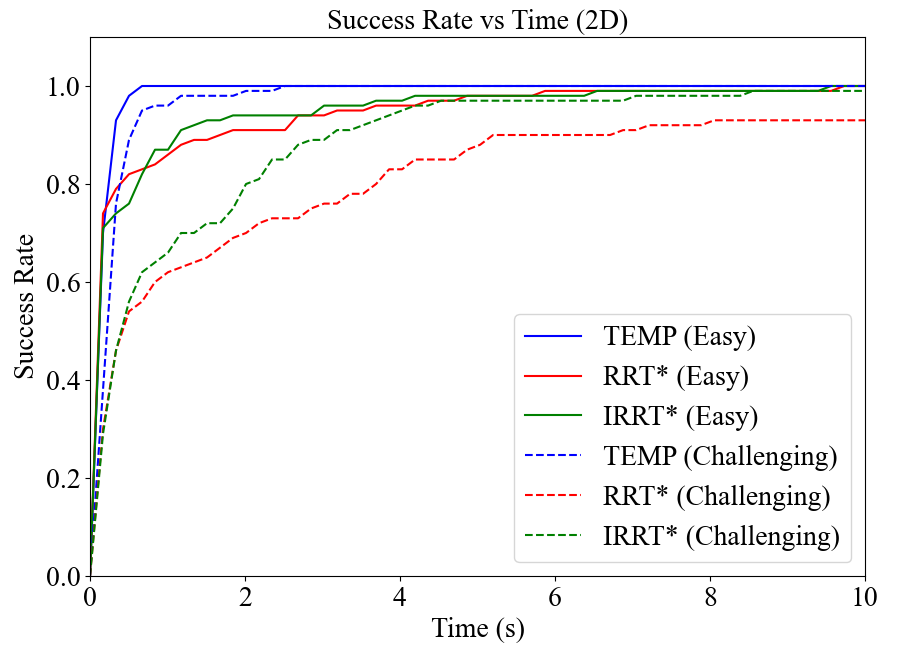}}
\hfil
\subfloat{\includegraphics[width=0.33\textwidth]{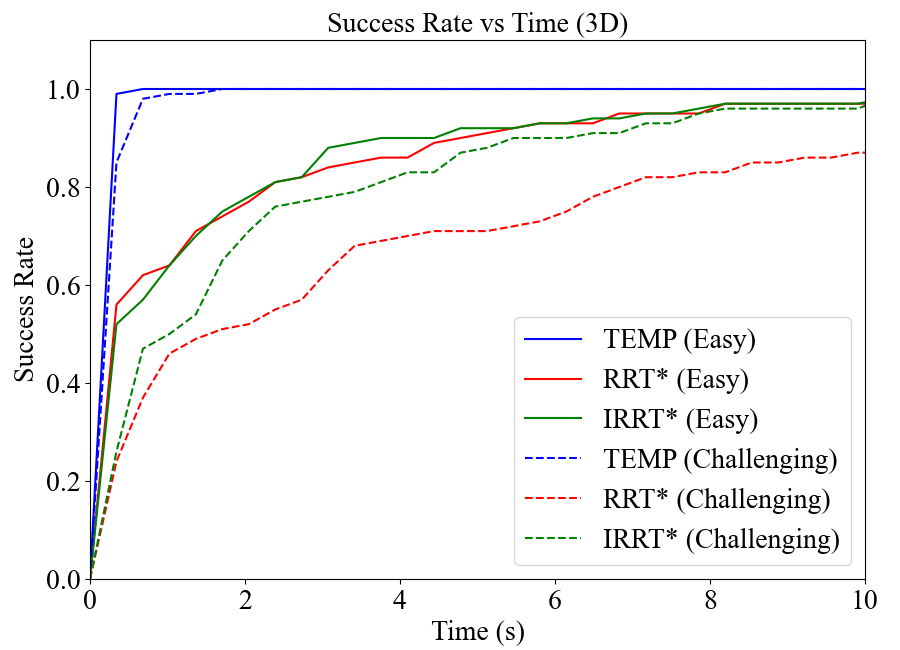}}
\hfil
\subfloat{\includegraphics[width=0.33\textwidth]{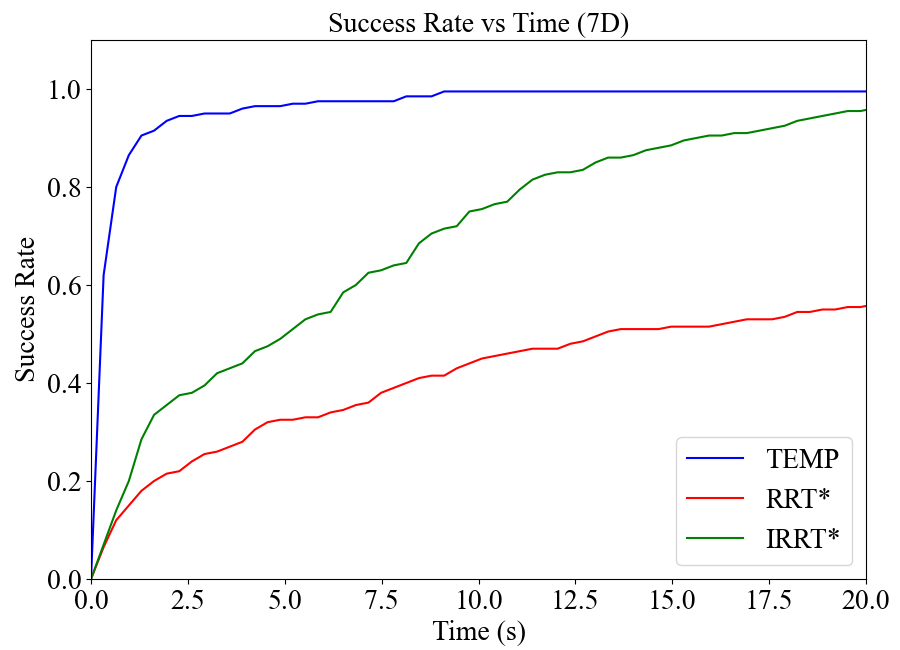}}
\caption{Planning success rate versus time curves for TEMP, RRT*, and IRRT* in 2D, 3D, and 7D scenarios.}
\label{Fig:SuccessRate}
\end{figure*}

\begin{table}[!t]
\caption{Comparison of Planning Time and Number of Nodes}
\label{Tab:ComTimeNodes}
\centering
\setlength\extrarowheight{2pt} 
\begin{tabular}{c c c c c c c c}
\hline
\multicolumn{2}{c}{Planning} & \multicolumn{2}{c}{\textbf{TEMP}} & \multicolumn{2}{c}{RRT*} & \multicolumn{2}{c}{IRRT*} \\
\cline{3-8} 
\multicolumn{2}{c}{Tasks} & \textbf{Time} & \textbf{Nodes} & Time & Nodes & Time & Nodes \\
\cline{1-8} 
\multirow{2}{*}{2D} & Easy & \textbf{0.154} & \textbf{117} & 0.575 & 322 & 0.531 & 274 \\
\cline{2-8} 
& Chall. & \textbf{0.283} & \textbf{211} & 2.165 & 675 & 1.197 & 604 \\
\cline{1-8} 
\multirow{2}{*}{3D} & Easy & \textbf{0.145} & \textbf{89} & 1.566 & 599 & 1.462 & 426 \\
\cline{2-8} 
& Chall. & \textbf{0.222} & \textbf{134} & 4.246 & 1172 & 2.085 & 705 \\
\cline{1-8} 
\multicolumn{2}{c}{7D Manip.} & \textbf{0.754} & \textbf{208} & 18.50 & 1501 & 6.940 & 782 \\
\hline
\end{tabular}
\end{table}

The success rate of planning within a time frame is an important indicator of an algorithm's stability. Fig. \ref{Fig:SuccessRate} demonstrates that, whether in planning for 2D, 3D, or 7D,  the TEMP's success rate approaches 1 more quickly compared to the RRT* and IRRT*. Moreover, with increasing planning complexity and dimensionality, the gap between TEMP and the other two algorithms widens, revealing that TEMP possesses greater robustness, especially in more demanding planning scenarios.

\subsection{Attention-Guided Node Sampling}

\begin{figure*}[!t]
\centering
\captionsetup[subfigure]{font=normal,labelfont={rm},textfont=rm}
\subfloat[]{\includegraphics[width=0.35\textwidth]{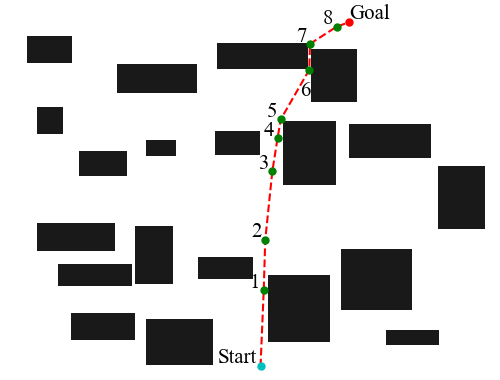}}
\hfil
\subfloat[]{\includegraphics[width=0.35\textwidth]{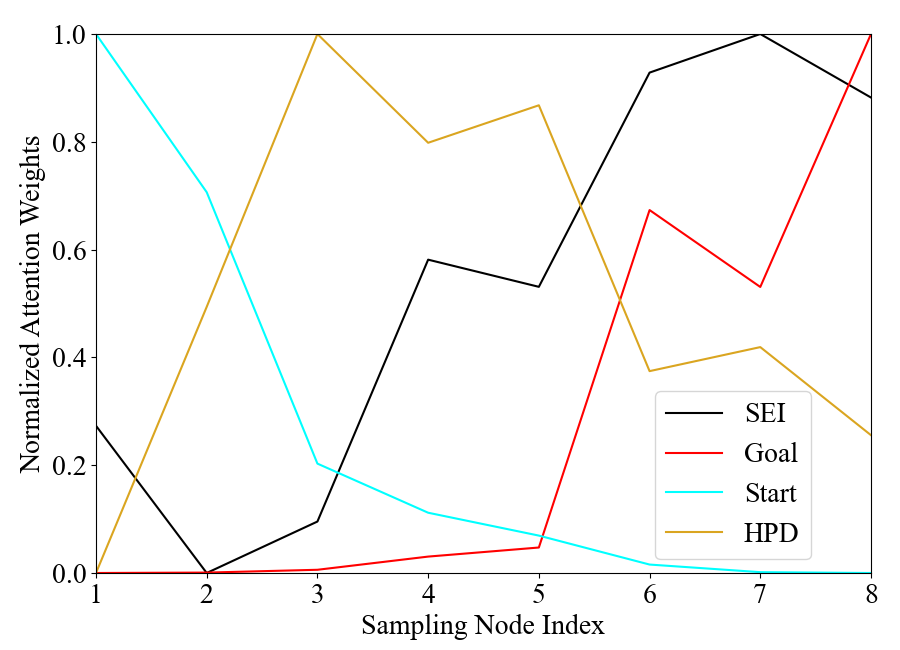}}
\caption{Typical sampling nodes and the corresponding attention weights allocations in a 2D planning task.}
\label{Fig:SamNodeandAtten}
\end{figure*}

\begin{figure*}[!t]
\centering
\captionsetup[subfigure]{font=normal,labelfont={rm},textfont=rm}
\subfloat[Scenario 1, TEMP]{\includegraphics[width=0.24\textwidth]{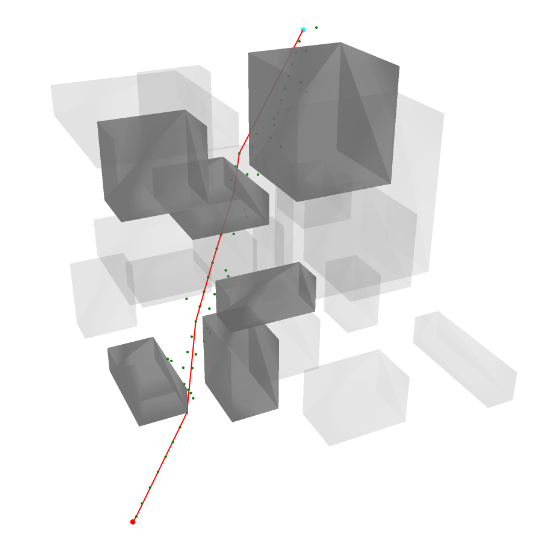}}
\hfil
\subfloat[Scenario 1, RRT*]{\includegraphics[width=0.24\textwidth]{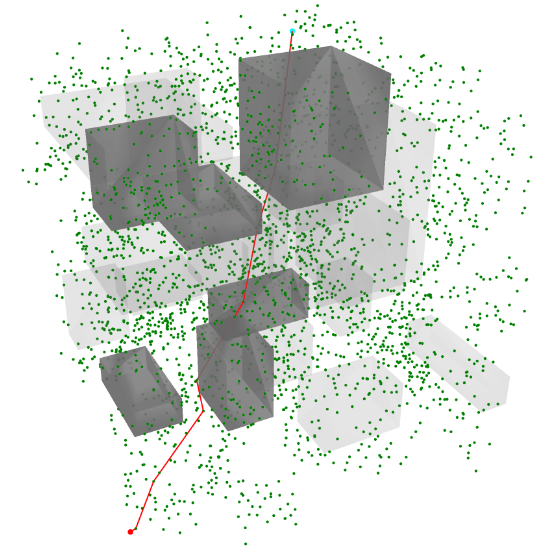}}
\hfil
\subfloat[Scenario 2, TEMP]{\includegraphics[width=0.24\textwidth]{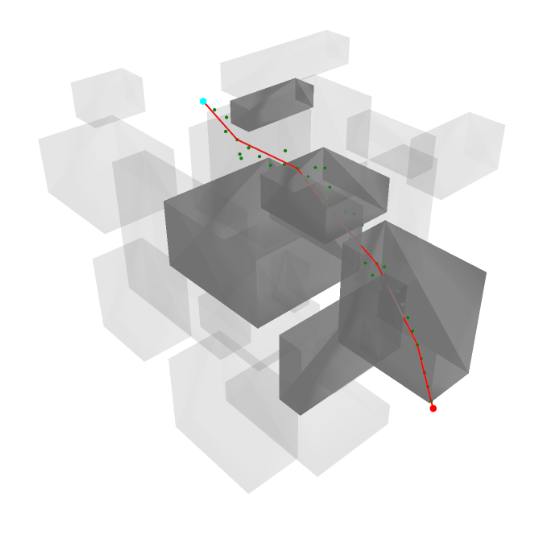}}
\hfil
\subfloat[Scenario 2, RRT*]{\includegraphics[width=0.24\textwidth]{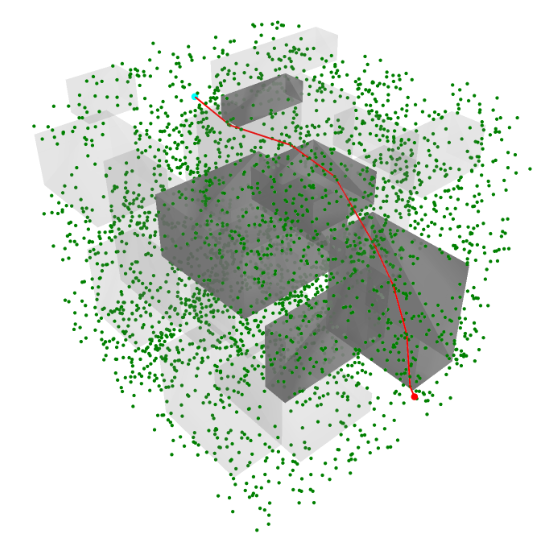}}
\caption{Performance assessment of TEMP versus RRT* in 3D planning, focusing on paths of comparable quality. (a) $t$ = 0.10 s, $N$ = 64, $\mathcal{J}$ = 28.33; (b) $t$ = 11.71 s, $N$ = 2683, $\mathcal{J}$ = 29.05; (c) $t$ = 0.07 s, $N$ = 42, $\mathcal{J}$ = 22.56; (d) $t$ = 12.32 s, $N$ = 3041, $\mathcal{J}$ = 23.34.}
\label{Fig:PlanTEMPvsRRTS}
\end{figure*}

We further explore how the attention mechanism within TEMP dynamically adjusts its focus across different information sources to guide the sampling process. Fig. \ref{Fig:SamNodeandAtten} presents a 2D planning case where part (a) illustrates typical sampling nodes from the start to the goal, and part (b) details the distribution of attention weights upon acquiring specific sampling nodes. To standardize the display scale of attention weights related to SEI and HPD, as well as the planning start and goal, we have applied min-max normalization, which facilitates an analysis of the relative contributions of each attention type on a comparable scale:
\begin{equation}
\label{EQ:AttenNorm}
\hat{\omega}_{j}^{k} = \frac{\omega_{j}^{k} - \min (\Omega^{k})}{\max (\Omega^{k}) - \min (\Omega^{k})}
\end{equation}
\noindent where ${{\Omega}^{k}}$ represents a specific set of attention weights for $k \in \{\text{SEI}, \text{Goal}, \text{Start}, \text{HPD}\}$, $j$ is the index of a sampling node, and $\omega_{j}^{k}$ and $\hat{\omega}_{j}^{k}$ represent the original and normalized attention weights, respectively. The attention weights $\omega$ are computed as:
\begin{equation}
\label{EQ:AttenWeights}
\omega = \frac{1}{h} \sum_{i=1}^h \mathrm{softmax}\left(\frac{Q_i K_i^T}{\sqrt{d_k}}\right)
\end{equation}
\noindent where $i$ is the index of the attention head and $h$ is the total number of heads. In dense and narrow environments, such as at nodes 6 and 7, obstacles heavily constrain path selection. TEMP enhances barrier perception by intensifying its focus on SEI. Conversely, in open areas like node 2, the model redirects its attention from SEI to HPD to improve path quality. At the outset of planning, TEMP strongly focuses on the start to promote effective exploration. As planning progresses, it gradually shifts focus towards the goal, increasing target-directedness. However, special circumstances exist, such as planning through a narrow passage at node 7, where overly goal-directed sampling could result in collisions; thus, compared to node 6, TEMP reduces its attention to the goal.

\begin{figure}[!t]
\centering
\includegraphics[width=0.45\textwidth]{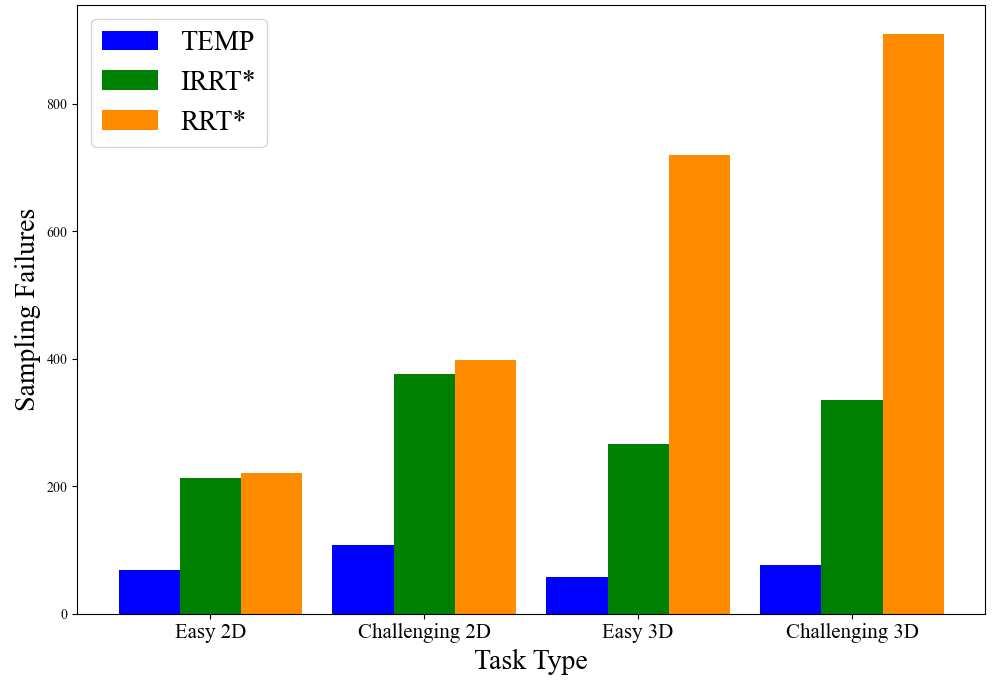}
\caption{Comparison of average sampling failures per planning task.}
\label{Fig:SamplingFailures}
\end{figure}

Fig. \ref{Fig:FigureinIntro} and Fig. \ref{Fig:PlanTEMPvsRRTS} illustrate how TEMP and RRT* generate sampling nodes to achieve paths of comparable quality. RRT* requires a significantly higher number of random samples to explore the solution space. In contrast, TEMP employs fewer nodes for high-quality paths through an efficient sampling strategy that concentrates nodes in critical areas. The average sampling failures for TEMP, RRT*, and IRRT* are depicted in Fig. \ref{Fig:SamplingFailures}. TEMP has significantly fewer failures than the others via attention-guided sampling, reducing reliance on collision detection and optimizing computational resource utilization.

\begin{figure*}[!t]
\centering
\subfloat{\includegraphics[width=0.33\textwidth]{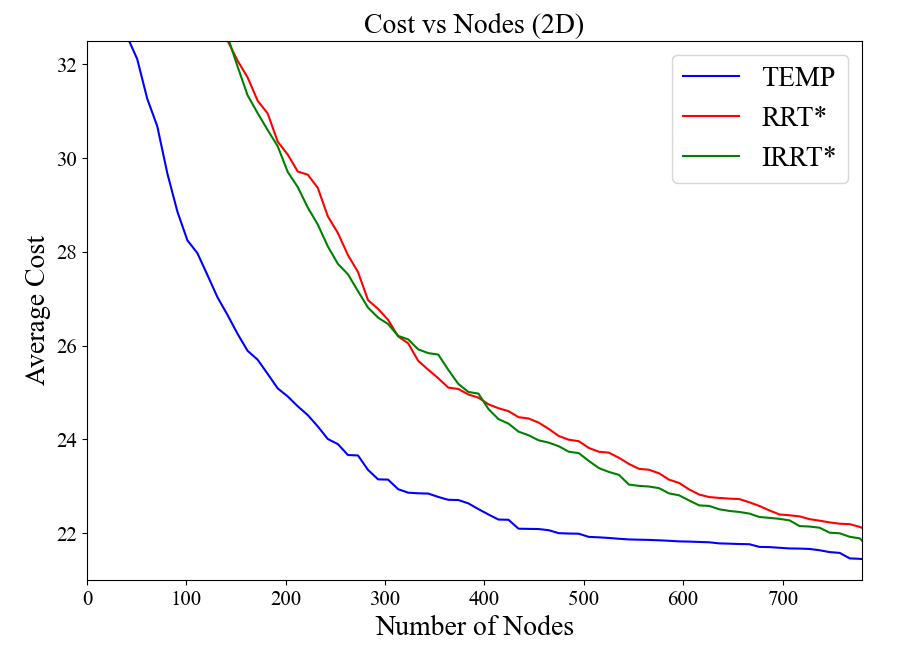}}
\hfil
\subfloat{\includegraphics[width=0.33\textwidth]{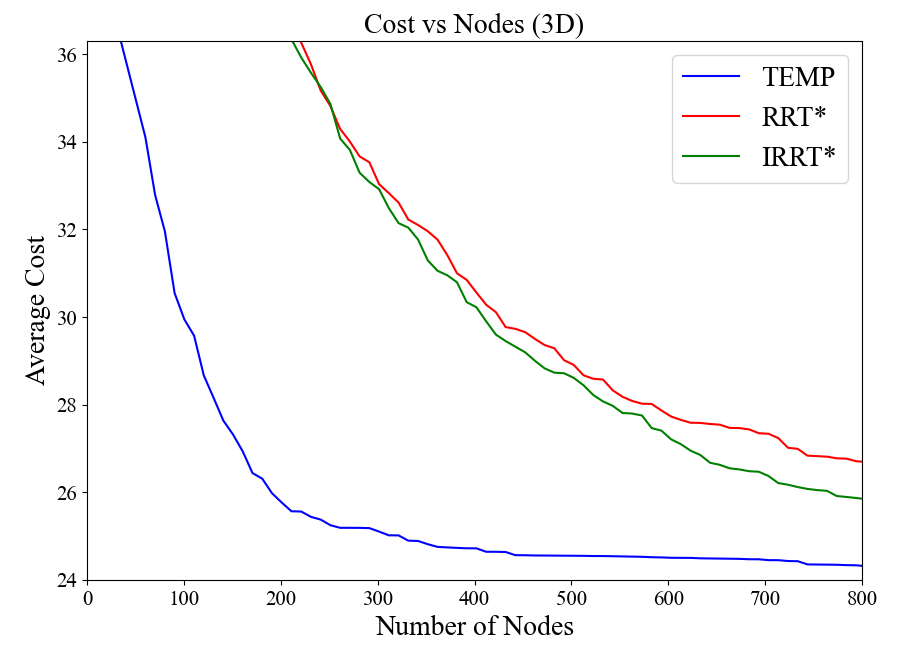}}
\hfil
\subfloat{\includegraphics[width=0.33\textwidth]{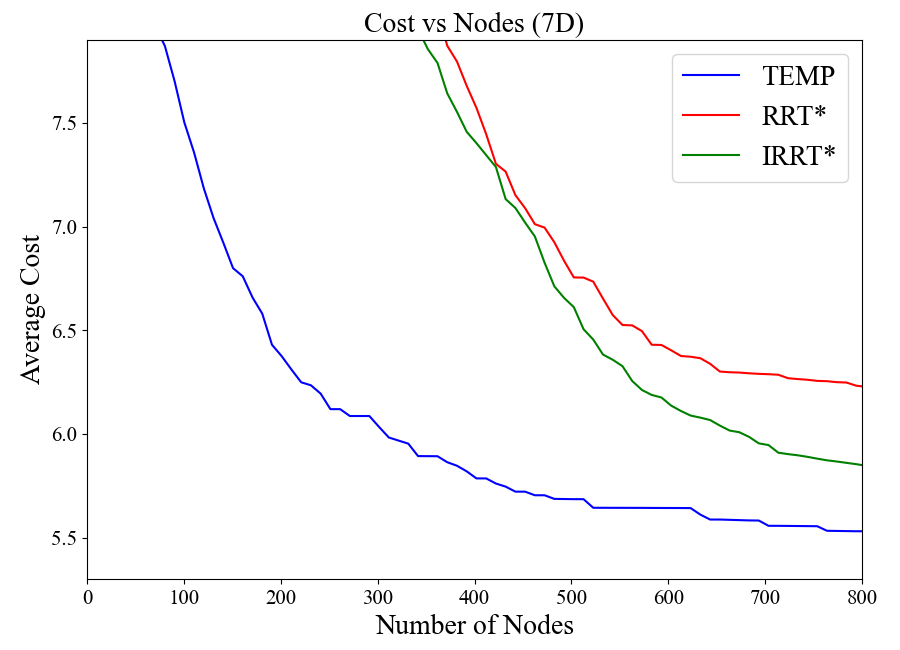}}
\caption{Comparative analysis of average path cost and number of nodes for TEMP, RRT*, and IRRT* in 2D, 3D, and 7D scenarios.}
\label{Fig:CostvsNodes}
\end{figure*}

The relationship between the average path cost and the number of nodes is illustrated in Fig. \ref{Fig:CostvsNodes}. When a planning attempt fails to find any feasible paths, the path cost is typically assigned an infinite value, complicating the computation of average path cost. Thus, for a not-yet-successful task, we define the cost as the maximum in that task multiplied by a penalty factor of 1.5. According to the data, TEMP achieves cost convergence with fewer nodes than RRT* and IRRT*, thereby suggesting that attention-guided sampling more effectively explores potential areas leading to lower-cost paths.

\section{Conclusions}

In this study, we propose the TEMP, a novel motion planning framework that combines EISE and MPT to overcome the challenges of path quality inconsistency and suboptimal search efficiency typical of SBMPs. The EISE converts environmental data into SEI, thereby providing semantically rich, structured representations. Subsequently, the MPT module tunes the focus on the SEI, task objectives, and HPD through the exploitation of attention mechanisms, resulting in more rational generation of sampling nodes. In comparison with advanced SBMPs, TEMP exhibits notable performance, particularly in challenging or high-dimensional motion planning.

In future research, we will evaluate the potential for extending the application boundaries of TEMP and attention mechanisms within the realm of motion planning, including multi-robot systems and motion dynamics constrained planning issues. Additionally, more concrete semantic encoding is crucial for the algorithm's deployment. Therefore, we plan to integrate supplementary intrinsic properties of obstacles into the semantic information to provide a broader understanding of the environment for the planning network.

\bibliographystyle{IEEEtran}
\bibliography{references}

\vspace{11pt}

\vfill

\end{document}